\documentclass[conference]{IEEEtran}
\usepackage{cite}
\usepackage{amsmath,amssymb,amsfonts}
\usepackage[linesnumbered,ruled]{algorithm2e}
\usepackage{graphicx}
\usepackage{textcomp}
\usepackage{xcolor}

\usepackage{array}
\newenvironment{conditions}
  {\par\vspace{\abovedisplayskip}\noindent\begin{tabular}{>{$}l<{$} @{${}={}$} l}}
  {\end{tabular}\par\vspace{\belowdisplayskip}}

\usepackage{subfiles}
\usepackage[font=scriptsize]{subfig}

\def\BibTeX{{\rm B\kern-.05em{\sc i\kern-.025em b}\kern-.08em
    T\kern-.1667em\lower.7ex\hbox{E}\kern-.125emX}}
\begin{document}

\title{Multi-agent Deep Reinforcement Learning for
Zero Energy Communities}

\author{\IEEEauthorblockN{Amit Prasad}
\IEEEauthorblockA{\textit{School of Computer Science and Statistics} \\
\textit{Trinity College Dublin}\\
Dublin, Ireland \\
prasada@tcd.ie}
\and
\IEEEauthorblockN{Ivana Dusparic}
\IEEEauthorblockA{\textit{School of Computer Science and Statistics} \\
\textit{Trinity College Dublin}\\
Dublin, Ireland \\
ivana.dusparic@scss.tcd.ie}
}

\maketitle

\begin{abstract}
A Zero Energy Building (ZEB) has its net energy usage over a period of one year as zero, i.e., its energy use is not larger than its overall renewables generation. A collection of such ZEBs forms a Zero Energy Community (ZEC). This paper addresses the problem of energy sharing in such a community. This is different from previously addressed energy sharing between buildings as our focus is on the improvement of community energy status, while traditionally research focused on reducing losses due to transmission and storage, or achieving economic gains. We model this problem in a multi-agent environment and propose a Deep Reinforcement Learning (DRL) based solution. Results indicate that with time buildings learn to collaborate and learn a policy comparable to the optimal policy, which in turn improves the ZEC's energy status. Buildings with no renewables preferred to request energy from their neighbours rather than from the supply grid.
\end{abstract}

\begin{IEEEkeywords}
Deep Reinforcement Learning, Energy sharing optimization, Zero Energy Community.
\end{IEEEkeywords}

\section{Introduction}
\label{sec:intro}
Advances in renewable energy generation, lower cost of the required hardware and increased storage capacity have made buildings or independent houses increasingly self-sustainable. If the net energy usage of such self-sustainable buildings over a period of one year is zero, they are called Zero Energy Buildings (ZEBs). However, achieving an exact net zero status is difficult, and hence the term \textit{nearly} ZEB (nZEB) is used in practice. The definition of nZEB states that, the net balance between export and import of energy over a period of time must be zero or even positive \cite{sartori2010criteria}. ZEBs are receiving increased attention in the recent years due to increasing demand and thus pressure on non-renewable sources of energy. The European Union(EU) has stated in its 2010 \cite{recast2010directive} recast that by the year 2020 all new buildings have to consume nearly zero energy. Similarly, the Japanese government too has set similar targets for the year 2030 \cite{kayo2014energy}. 

This concept of nZEB, when extended to a group of buildings, is called a \textit{nearly Zero Energy Community (nZEC)}. \cite{lopes2016cooperative} introduces the concept of cooperative ZEC (CNet-ZEC) and defines a ZEC as a collection of \textit{only} ZEBs having its annual energy balance as zero. However, this might not be accurate. The definition above describes a \textit{ZEB community}, as opposed to a ZEC. We maintain a subtle distinction between a ZEB community and a ZEC. The former is a community in which \textit{all} buildings satisfy the nZEB status whereas in the latter some buildings might not, but as a group they have net zero energy balance. All nZEB communities are nZECs but the inverse might not be true. We redefine nZEC formally as, \textit{"A micro-grid that has distributed generation, storage, delivery, consumption and nearly zero net annual energy balance."}

There is high uncertainty of energy production in an nZEC due to distributed generation that causes some buildings to produce insufficient energy with respect to their energy demand. Such buildings then request for additional energy from the supply grid at a higher cost. As an alternative to this, buildings can request for additional energy from the neighbouring buildings. This reduces the losses due to transmission and promotes green energy.  Moreover, buildings with surplus energy may also benefit economically by sharing with neighbours. In this work, we address the problem of energy sharing in such an energy community. We model this community as a multi-agent environment where each agent represents a building. Previous works \cite{mocanu2018line} have demonstrated that deep reinforcement learning (DRL) is an effective technique for energy management in a single building management system. Considering this, we propose a DRL-based solution to optimize energy sharing between multiple such buildings. Intelligent agents representing buildings, learn over time the appropriate behaviour to share energy in order to achieve a nearly zero energy status as a community.

The rest of the paper is organized as follows. We firstly discuss the existing work done in energy sharing between buildings in section \ref{related-work}. In section \ref{design-and-implementation} we present the design and implementation of our proposed intelligent energy sharing solution. In section \ref{sec:eval-results} we evaluate and present the results of our solution. Finally, in section \ref{conclusion-and-fw} we conclude the paper and discuss the issues that remain open along with avenues for future work.
\section{Related Work}
\label{related-work}

Energy can be more efficiently utilized if the excess energy produced on-site can be shared with other nearby homes that require it. This can benefit the seller economically by reducing the transmission losses that would have occurred otherwise if energy was requested from the supply grid, or conversion losses if it was stored in batteries. To enable this, various strategies for energy sharing amongst homes exist.

The authors of \cite{olivares2011centralized} define Central Energy Sharing (CES) systems as those that consists of a central controller responsible for managing all distributed energy resources in the micro-grid as well as the forecasting systems, and schedules resources accordingly. This type of system allows broad observability but reduces the flexibility. Work in \cite{zhu2013sharing} introduces a CES system that classifies homes into sets of suppliers and consumers, and then uses a greedy strategy to initiate energy exchange between homes. Another interesting system - IDES \cite{Zhong2015} uses a sophisticated distributed energy generation and sharing approach (DES) with a novel pricing model to incentivize energy sharing between homes. 

There have been other works that focus on energy sharing between buildings but no direct work has been done on intra-nZEC energy sharing for the community benefit. Previous works on only energy sharing between buildings focus on economic gains \cite{stanczak2015dynamic,Zhong2015}, optimizing transactions \cite{swaminathan2018investigations} and thereby reducing transmission and storage loses at the individual building level. \cite{lopes2016cooperative} defines nZEC as a collection of \textit{only} ZEBs having its annual energy balance as zero.  However, \cite{kayo2014energy} argues that achieving a ZEB status without a grid is very difficult. The authors propose a solution by defining \textit{energy communities} using a basic energy matching algorithm and conclude that energy sharing can help achieve ZEB status for individual buildings. Additionally, in the previous section, we have argued that in a ZEC buildings may have varying generation capacities and some may even have none.

Previous related works in energy sharing, energy and cost optimization have used techniques  like linear programming, dynamic programming, heuristic methods such as  particle swarm optimization, game theory, and so on \cite{Zhong2015,zhu2013sharing}. The general problem with majority of the algorithms is that, for optimization they compute partial or the entire solution space to choose the best one, and hence are time consuming. \cite{mocanu2018line} explores an interesting approach that avoids computing the entire search space using \textit{Deep Reinforcement Learning (DRL)} to optimize schedules for building's energy management systems. The authors investigate two DRL based algorithms, Deep Policy Gradient (DPG) and Deep Q-Learning (DQN) for building a on-line large scale solution and conclude it to be more effective than traditional solutions. \cite{franccois2016deep} uses DRL to model a solution that optimizes activation of energy storage devices considering the uncertainty of energy generation and consumption. \cite{nguyen2017deep} uses DPG to optimize the schedule of energy consuming devices in a dynamic energy pricing environment and reports great results.

Considering the success of DRL based techniques in single-house scenarios and also its effectiveness in building on-line large scale solutions, we propose a DRL-based energy sharing solution to address the issue of energy sharing between buildings.
\section{Design of DRL-based nZEC}
\label{design-and-implementation}

We propose modelling an nZEC community as a multi-agent environment, where each agent represents a building. Every agent learns the optimal behaviour independently and is entirely responsible for making energy transactions on behalf of that building. We identify two main components that are central to building a solution for this - the \textit{DRL agent}, which learns the behaviour of an individual household and a \textit{Community Monitoring Service (CMS)} to enable collaboration between the agents. 
In the following sections we present the details of each.

\subsection{DRL-based Energy Management Agent}

Reinforcement Learning (RL) is a machine learning approach that enables intelligent agents to learn the optimal behavior via trail-and-error \cite{sutton1998introduction}. RL is particularly suited for the implementation of self-organizing behaviours in large scale systems as it does not require a predefined model of the environment \cite{dusparic2017residential}. Q-learning is one such model-free RL algorithm that allows intelligent agents to learn to associate actions with expected long term rewards of taking that action in a particular state \cite{watkins1989learning}. DRL is an extension of the traditional RL algorithm, and uses Neural Networks (NN) at its core to discover non linear solutions. This paper uses a DRL based algorithm called DQN to approximate Q-values. A Q-value is the expected value of taking an action in a particular state while considering the expected long term reward of taking that action in that state. 

We use the following network configuration after experimenting with various models --
\begin{itemize}
    \item \textbf{Input Layer} with 63 neurons representing the encoded \textit{State}, \textit{Action} and \textit{Reward}.
    \item \textbf{2 x Hidden Layers} with 100 neurons each and the activation function as sigmoid. 
    \item \textbf{1 Output neuron} representing the Q-value with linear activation function.
\end{itemize}

We use a popular optimization technique called \textit{Stochastic Gradient Descent (SGD)} to refine the NN, and is suitable for systems that employ that employ machine learning on a large scale \cite{bottou2010large}. Additionally, to reduce correlation between the data used (i.e. state-action pair and reward received) to train the NN and also to overcome issues with convergence we have used a technique called \textit{Combined Experience Replay} \cite{zhang2017deeper} that adds the latest experience to the mini-batch of random experiences.

\begin{algorithm}
\DontPrintSemicolon
\caption{Agent Learning Algorithm}
\label{alg:1}
\SetAlgoLined
env $\gets$ Environment() \tcc*[r]{env is an instance of the environment}

\While{true} {
consumption $\gets$ env.preceptEnergyConsumption()\;
generation $\gets$ env.perceptEnergyGeneration()\;
state $\gets$ updateEnergyBalance(consumption, generation)\;
legalActions $\gets$ getLegalActions()\;
chance $\gets$ generateRandomNumber()\;

\eIf{chance $\leq \epsilon$}
{
action $\gets$ chooseRandomAction(legalActions)\;
}{
action $\gets$ selectBestActionFromPolicy(legalActions)\;
}

nextState, reward $\gets$ env.takeAction(action)\;

updateAgentLearning(state, action, reward)\;

}
\end{algorithm}

The overall learning process of a DRL agent is summarized in the Algorithm \ref{alg:1}. An agent senses the environment conditions (energy consumption and generation) and translates them into a state vector, which is then processed to select a suitable action. We describe available states and actions below. 

\subsubsection{\textbf{State space}}
Every agent has a \textit{State} (line 5 - Algorithm \ref{alg:1}) object that has \textit{AgentState} and \textit{EnvironmentState} objects embedded in it.  This helps an agent to keep track of its own internal state and its perspective of the world. The AgentState object keeps track of the energy consumed, generated and stored by that agent at a particular time instant. Similarly, the EnvironmentState object keeps track of the energy balance of the community. This state information along with time of the day (discretized into 48 intervals of half hour each), and day of the week is encoded and fed into the NN for training. The continuous energy values in the State object are also discretized to values 0, 1, 2 and 3 representing none, low, medium and high energy states respectively before they are fed into the NN.
%
%
%

\subsubsection{\textbf{Action set}}  
In the simplest scenario, in a energy sharing environment an agent representing a energy user faces the following choices to manage its energy requirements --

\begin{itemize}
    \item Consume and store excess energy.
    \item Request neighbour for additional energy.
    \item Request supply grid for additional energy.
    \item Grant energy request from a neighbour.
    \item Deny energy request from a neighbour.
\end{itemize}



These sets of actions along with the encoded state information and reward is passed to the \textit{DRL engine} that returns an appropriate action whenever requested for. The reward function generates a reward for taking an action in a particular state based on the feedback from the environment (line 13 - Algorithm \ref{alg:1}). As agents in our system have a common goal to achieve zero energy status as a community, they receive similar rewards based some global information discussed in the next section.

\subsection{Community Monitoring Service}

Literature in cooperative strategies \cite{claus1998dynamics,riedmiller2000karlsruhe,stone2005reinforcement} suggests the use of shared rewards or global rewards to enable cooperation between individual learners.

\subsubsection{\textbf{Reward model}}  

In the simplest form, a negative of the community energy status can used as a global reward.

\begin{equation}
reward = - \bigg(\sum_{i=1}^{n}c(h_{i}) - g(h_{i})\bigg)
\end{equation}
where:
\begin{conditions}
c(h_{i}) & energy consumed by the $i^{th}$ house \\
g(h_{i}) &  energy generated by the $i^{th}$ house 
\end{conditions}

To enable this, we introduce a \textit{Community Monitoring Service (CMS)}. The CMS acts as a agent group membership management service with functionalities like agent joining the group, agent leaving the group and, maintaining a list of active agents. Apart from this, CMS also collects individual energy status' from all agents at regular intervals and calculates the community energy status. Intelligent agents can access this information via HTTP calls and use it to calculate their rewards based on the action taken.

\subsection{Hyper-parameter Tuning}

All agents are trained in an episodic manner and, are rewarded at the end of each episode. We experimented with various values for the learning rate($\alpha$) and for further experiments selected $0.125*10^{-3}$ that led to convergence. Similarly, we choose the value of discount factor($\gamma$) to be 0.99 after experimenting with a range of values between 0.5 and 0.99. 

An RL agent needs to decide between exploring new actions that might lead to new better or worse states and exploiting already known best actions. This is handled using $\epsilon-greedy$ action selection policy. An agent chooses a exploratory action with probability $\epsilon$ and exploits its existing knowledge of best known actions at all other times. Although advanced techniques are available, in this design we simply decay the value of $\epsilon$ by 0.8 ($new\_\epsilon = \epsilon*0.8$) after every $(1/10_{th})* number\_of\_training\_episodes$ starting with initial value as 1 and eventually set this value to 0 in complete exploitation mode.


The energy consumption dataset used in this simulation was generated using Load Profile Generator\footnote{https://www.loadprofilege-nerator.de/}, that models the behaviour of the people living in a house to generate consumption data. We have used the energy consumption values generated every half an hour during 3 weekdays for this simulation. This allows every agent to experience a minimum of 144 states in each training episode. The number states experienced also depends on the number on interactions an agent has with other agents for energy sharing.  We assume that every house may or may not be equipped with solar panels, but is the only source of on-site energy generation. For this, we have used NREL's NSRDB\footnote{https://nsrdb.nrel.gov/nsrdb-viewer} solar exposure dataset of Toronto City.
\section{Evaluation and Results}
\label{sec:eval-results}

\subsection{Experimental Setup and Parameters}
\label{subsec:exp-setup}
To simulate different generation and demand patterns, we have evaluated our approach in 2 sets of weather conditions- Winter and Summer. During winter, House 1, House 2 and House 3 have average daily consumption as 11.01, 9.49, and 10.03 $kWh$ respectively. Similarly, during summer their average daily consumption is 12.12, 11.68, and 8.27 $kWh$ respectively. House 4 uses the same dataset as House 1 and therefore has the same energy consumption profile as House 1. As all these houses are part of the same community, we assume that they are present in the same geographical region and therefore experience similar solar exposure. The average daily solar exposure during winter is 11770 $W/m^{2}$ and during summer is 18850 $W/m^{2}$.

Further, in each set of conditions, we have evaluated 3 different scenarios representing different community configurations (i.e., combinations of initial stored energy value and the numbers of solar cells available to a household) and 3 different scales (i.e. different number of households). In all evaluation scenarios agents were trained for 3 simulated days, for 500 episodes. 

We begin training our agents at 00:00 hrs i.e. at midnight and as there is no solar exposure during that time all agents are forced to borrow energy from the supply grid. This introduces a bias that leads to problems in discovering  the optimal policy. To overcome this bias, we provide agents with an initial battery charge. The batteries used in this simulation are have a maximum power voltage of 12V and a charge rate of 100 amps for 20 hours. We convert this into $kWh$ for ease of calculation (calculates to 1.2 $kWh$). Each house is equipped with 6 such batteries and has a total storage capacity of (1.2 * 6 =) 7.2 $kWh$.

We have tested for the following scenarios:

\begin{itemize}
\item Scenario 1 - 3 houses having varying generation capacity and initial battery charge
\item Scenario 2 - 4 houses having varying generation capacity and initial battery charge (with one of the houses having 0 generation capacity)
\item Scenario 3 - 10 houses having varying generation capacity and initial battery charge
\end{itemize}

\noindent Configuration for the houses in \textit{Scenario 1} and  \textit{Scenario 2} is described in Table \ref{tab:eval1} below:

\begin{table}[h!]
    \centering
    \begin{tabular}{|c|c|c|c|}
        \hline
        \textbf{House No.} & \textbf{Agent}  & \textbf{n*(SCells)} & \textbf{Batt. Init (kWh)}\\
        \hline
        1 & Alice & 72 & 7.2\\
        2 & Bob & 54 & 2.5\\
        3 & Charlie & 12 & 5.0\\
        \hline
        \hline
        4 & Dave & 0 & 0\\
        \hline
    \end{tabular}
    \setlength{\belowcaptionskip}{-15pt}
    \caption{Configuration for 3 and 4 houses having varying generation capacity.}
    \label{tab:eval1}
\end{table}

\subsection{Results and Analysis}
\textit{Always Share}, \textit{Random}, and \textit{Never Share} labels in figures \ref{fig:sub1-3a-winter}, \ref{fig:sub4-3a-winter-cg} represent the baseline strategies: always-share-energy strategy, random-action-selection strategy, and no-energy-sharing strategy. In these scenarios, the optimal strategy to achieve ZEC status is to always share leftover energy with neighbours as there are no other components that will affect the decisions taken by agents. However, evaluating whether agents are capable of learning that to be the optimal strategy (versus not sharing, or charging local batteries), and to what extent, will enable integration of this approach into more complex energy balancing scenarios that include, for example, dynamic pricing models.


\subsubsection{\textbf{Scenario 1}}
\begin{figure}
\centering
\includegraphics[width=0.9\linewidth]{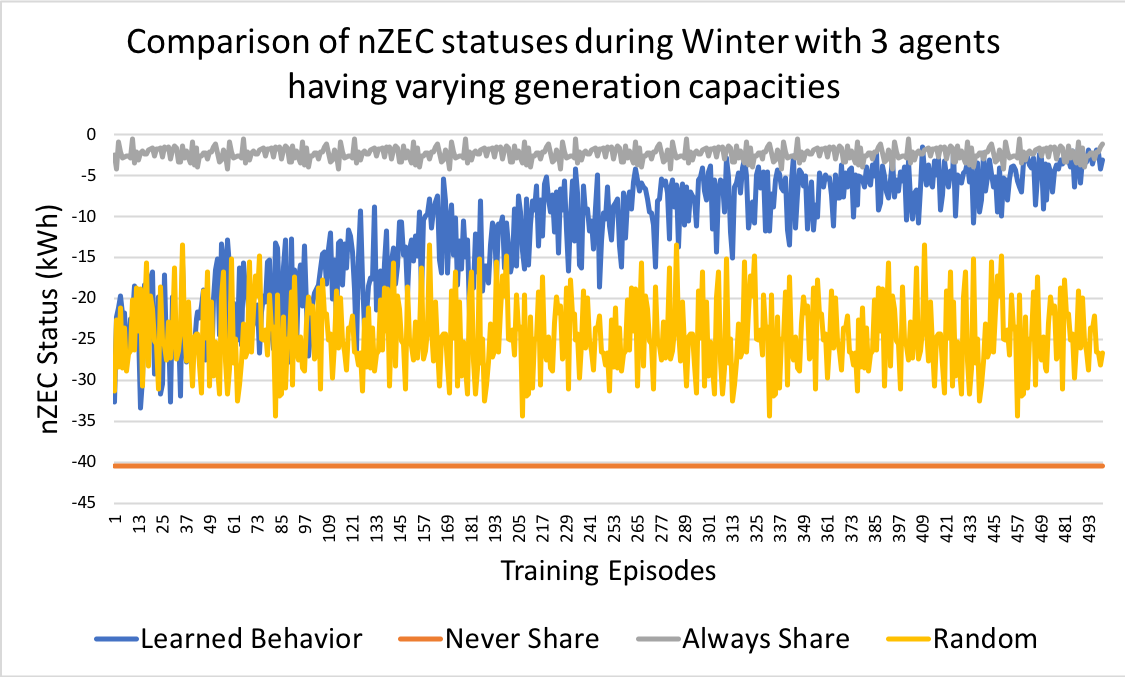}
\setlength{\belowcaptionskip}{-15pt}
\caption{Scenario 1 - Comparison of different strategies during Winter}
\label{fig:sub1-3a-winter}
\end{figure}

\begin{figure}
\includegraphics[width=0.9\linewidth]{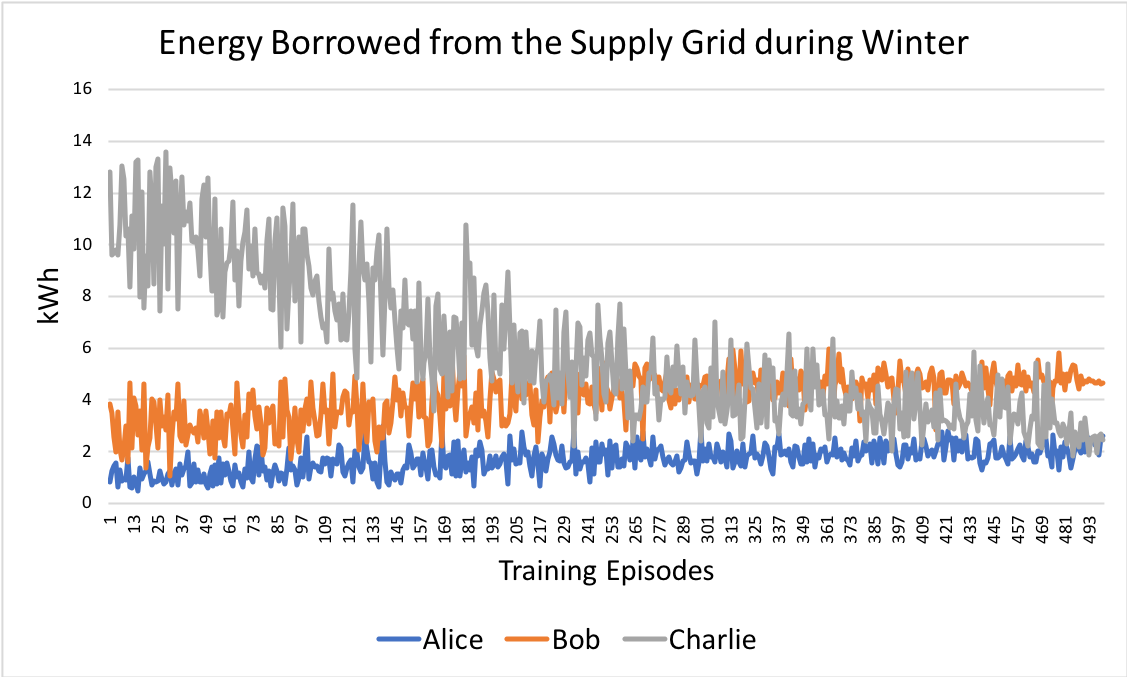}
\setlength{\belowcaptionskip}{-15pt}
\caption{Scenario 1 - Energy borrowed from the Supply Grid during Winter}
\label{fig:sub4-3a-winter-cg}
\end{figure}

During Winter, as agents train, they learn the optimal behaviour (indicated by the \textit{Learned Behaviour} label in fig. \ref{fig:sub1-3a-winter}) to always share energy. To achieve this optimal behavior agents learn to borrow more energy from their neighbours and less energy from the supply grid(fig. \ref{fig:sub4-3a-winter-cg}). However, we also observe that (fig. \ref{fig:sub4-3a-winter-cg}) 2 agents (\textit{Alice} and \textit{Bob}) learn to request for additional energy to the supply grid. This is because agents learn to operate selflessly in order to support the third agent (\textit{Charlie}) that has lesser renewable generation. This helps to improve the net zero energy balance of the community. Additionally, as houses share energy with each other, a deficit is introduced in their locally available batteries and they can now store the generated surplus energy which would otherwise have been wasted due to fully charged batteries.

During Summer, with the configuration in Table \ref{tab:eval1}, houses are entirely self-sustainable and therefore have no need to share energy with each other. As there is no energy sharing involved, 
all strategies exhibit the same nZEC status of -1.72 $kWh$. This minor negative status of 1.72 $kWh$ is due to the initial bias introduced by the training data (discussed in sec. \ref{subsec:exp-setup}).

\subsubsection{\textbf{Scenario 2}}
With the same setup as in Scenario 1, we introduce a fourth house with no source of renewable energy generation in Scenario 2. We have argued that such energy communities can exist (section \ref{sec:intro}) and hence the need for energy sharing in such communities. During Summer, agents were still able to learn a policy comparable to the optimal policy with a difference of only 7 $kWh$ (where nZEC status of optimal policy is 20 $kWh$) in their nZEC statuses. When compared with the no-energy-sharing strategy the agents perform very well (difference of ~64 $kWh$). Similar behaviour was observed during Winter too.


\subsubsection{\textbf{Scenario 3}}
In this scenario, we evaluate the ability of our system to learn the optimal solution when the number of agents grow. For this, we have considered 10 agents with varying generation capacities and initial charge. In both Seasons, our system performs better than a random-action-selection strategy and a no-energy-sharing strategy. During Winter, the distinction between \textit{Always Share} , \textit{Random}, and \textit{Learned Behaviour} curves is very minor, however, our solution still tends towards the optimal strategy.

%
%
%
%
%
%
\section{Conclusions and Future Work}
\label{conclusion-and-fw}

This paper introduces a new definition of nZEC that encompasses a mixture of buildings with varying levels of energy generation building upon the previous definition of nZEB. It also introduces a multi-agent DRL based solution for energy sharing between houses in such an nZEC community. Results indicate that DRL is a suitable technique for building intelligent agents that are able to collaborate with each other to optimize energy transactions between buildings and achieve net zero energy balance as a community. This behaviour allows us to build a nZEC with varying levels of distributed generation. Evaluations also indicate that our solution improves the nZEC status drastically when compared to a no-energy-sharing strategy and random-action-selection strategy. An improvement of 40 $kWh$ with 3 houses during winter and over 60 $kWh$ with 4 houses during summer over 3 days in the overall community's energy balance was found when compared to a no-energy-sharing strategy. Additionally, as an indirect effect of energy sharing, houses were able to produce more energy that consequently increased the flow of energy generated from renewable sources in the community.

We have trained the agents in an episodic manner and it would be interesting to observe their behaviour if they were trained in an on-line fashion. Further, we were not able to test this system on scale with hundreds or thousands of houses due to the limitation on physical resources we had and other issues related to the implementation and framework. Our design is simple and the focus is mainly on motivating agents to share energy without any energy pricing scheme involved. However, in real world systems this is not true and such integration is necessary, and is a potential future research. 

\section{Acknowledgement}
This publication is supported in part by research grants from Science Foundation Ireland (SFI) under Grant Numbers 13/RC/2077 and 16/SP/3804.
%
%

\end{document}